\documentstyle[colacl]{article}
\author{\hspace*{-0.5cm}Masaki Murata \hspace{0.6cm}  Hitoshi Isahara \\
\hspace*{-0.5cm}Communications Research Laboratory\\
\hspace*{-0.5cm}588-2, Iwaoka, Nishi-ku, Kobe, 651-2401, Japan\\
\hspace*{-0.5cm}{\tt \{murata,isahara\}@crl.go.jp}\\
\hspace*{-0.5cm}TEL: +81-78-969-2181, FAX: +81-78-969-2189
\And  \hspace*{2.5cm}Makoto Nagao \\
\hspace*{2.5cm}Kyoto University\\
\hspace*{2.5cm}Sakyo, Kyoto 606-8501, Japan \\
\hspace*{2.5cm}{\tt nagao@pine.kuee.kyoto-u.ac.jp}}
\title{\mbox{Pronoun Resolution in Japanese Sentences} 
\mbox{Using Surface Expressions and Examples}}
\begin{document}
\maketitle
\bibliographystyle{acl}

\small


\newenvironment{indention}[1]{\footnotesize \par\begingroup\addtolength{\leftskip}{#1}}{\par\endgroup}

\begin{abstract}
In this paper, we present a method of estimating 
referents of demonstrative pronouns, personal pronouns, 
and zero pronouns in Japanese sentences
using examples, surface expressions, topics and foci.
Unlike conventional work 
which was semantic markers 
for semantic constraints, 
we used examples for semantic constraints 
and showed in our experiments that 
examples are as useful as semantic markers. 
We also propose many new methods for estimating referents of pronouns. 
For example, we use the form ``X of Y'' for 
estimating referents of demonstrative adjectives. 
In addition to our new methods, 
we used many conventional methods. 
As a result, experiments using these methods 
obtained a precision rate of 87\% 
in estimating 
referents of demonstrative pronouns, personal pronouns, 
and zero pronouns 
for training sentences, 
and obtained a precision rate of 78\% 
for test sentences. 
\end{abstract}

\section{\normalsize \bf Overview}
\label{sec:deno_overview}

This paper describes how to resolve the referents of pronouns: 
demonstrative pronouns, personal pronouns, and zero pronouns. 
Pronoun resolution is especially important 
for machine translation. 
For example, if the system cannot resolve zero pronouns\footnote{
Omitted noun phrases are called {\it zero pronouns}.}, 
it cannot translate sentences containing them from Japanese into English. 
When the word order of sentences is changed and 
the pronominalized words are changed in translation into English, 
the system must detect the referents of the pronouns. 

A lot of work has been done 
in Japanese pronoun resolution \, 
\cite{kameyama1} \,\,
\cite{yamamura92_ieice} \,\, \cite{walker2} \,\, \cite{takada1} \, 
\cite{nakaiwa}. 
The main distinguishing features of our work are as follows: 

\begin{itemize}
\item 
In conventional pronoun resolution methods, 
semantic markers have been used for semantic constraints. 
On the other hand, we use examples for semantic constraints 
and show in our experiments that 
examples are as useful as semantic markers. 
This is an important result 
because the cost of constructing the case frame 
using semantic markers is generally higher than 
the cost of constructing the case frame 
using examples. 

\item 
We use examples in the form ``X {\it no} Y'' (Y of X) for 
estimating referents of demonstrative adjectives. 

\item 
We deal with the case 
when a demonstrative refers to 
elements that appear later. 

\item 
We resolve a personal pronoun in a quotation 
by determining who is the speaker and who is the listener. 
\end{itemize}

In this work, 
we used almost all the potentials of conventional methods 
and also propose a new method. 

\section{\normalsize \bf The Framework for Estimating the Referent}
\label{5c_wakugumi}

Prior to the pronoun resolution process, 
sentences are transformed into a case structure 
by a case structure analyzer \cite{csan2_ieice}. 
The antecedents of pronouns are 
determined by heuristic rules 
from left to right. 
Using these rules, 
our system assigns points to possible antecedents, and judges that 
the one having the maximum total score 
is the desired antecedent. 

\begin{figure}[t]
\vspace*{-0.8cm}
  \footnotesize
  \leavevmode
  \begin{center}
\fbox{
    \begin{minipage}[c]{7.5cm}
      \hspace*{0.7cm}Condition $\Rightarrow$ \{Proposal Proposal ..\}\\
      \hspace*{0.7cm}Proposal := (Possible-Antecedent \, Points)

\vspace*{-0.2cm}
    \caption{Form of {\sf Candidate enumerating rule}}
    \label{fig:kouho_rekkyo}
    \end{minipage}
}
  \end{center}
\end{figure}

\begin{figure}[t]
\vspace{-1cm}
  \footnotesize
  \leavevmode
  \begin{center}
\fbox{
    \begin{minipage}[c]{7.5cm}
      \hspace*{1.5cm}Condition $\Rightarrow$ (Points)
\vspace*{-0.2cm}
    \caption{Form of {\sf Candidate judging rule}}
    \label{fig:kouho_hantei}
    \end{minipage}
}
  \end{center}
\end{figure}

\begin{table*}[t]
\vspace*{-0.5cm}
  \footnotesize
  \begin{center}
  \caption{The weight in the case of topic}
  \label{tab:5c_topic}
    \leavevmode
\begin{tabular}[c]{|l|l|r|}\hline
\multicolumn{1}{|c|}{Surface expression}  & \multicolumn{1}{|c|}{Example}   & Weight \\\hline
Pronoun/zero-pronoun {\it ga}/{\it wa}&  (\underline{John} {\it ga} (subject)) {\it shita} (done).&21 \\\hline
Noun {\it wa}/{\it niwa}        &  \underline{John}{\it wa} (subject) {\it shita} (do).  &20 \\\hline
\end{tabular}
  \end{center}
\end{table*}

\begin{table*}[t]
\vspace*{-1cm}
  \footnotesize
  \begin{center}
  \caption{The weight in the case of focus}
  \label{tab:5c_focus}
    \leavevmode
\begin{tabular}[c]{|l|l|r|}\hline
\multicolumn{1}{|p{5cm}|}{Surface expression }  & \multicolumn{1}{|c|}{Example}   & Weight \\\hline
Pronoun/zero-pronoun {\it wo} (object)/ {\it ni} (to)/{\it kara} (from) & (\underline{John} {\it ni} (to)) {\it shita} (done).& 16 \\\hline
Noun {\it ga} (subject)/{\it mo}/{\it da}/{\it nara} & \underline{John} {\it ga} (subject) {\it shita} (do).  & 15 \\\hline
Noun {\it wo} (object)/{\it ni}/, /.       & \underline{John} {\it ni} (object) {\it shita} (do).  & 14 \\\hline
Noun {\it he} (to)/{\it de} (in)/{\it kara} (from)   & \underline{{\it gakkou} (school)} {\it he} (to) {\it iku} (go).  & 13 \\[0.1cm]\hline
\end{tabular}
  \end{center}
\end{table*}

Heuristic rules are classified into 
two kinds: 
{\sf Candidate enumerating rule}s 
and {\sf Candidate judging rule}s. 
{\sf Candidate enumerating rule}s 
are used in enumerating  candidate antecedents and 
giving them points (which represent the plausibility 
of being the correct antecedent). 
{\sf Candidate judging rule}s 
are used in 
giving points to the candidate antecedents 
selected by {\sf Candidate enumerating rule}s. 
These rules are shown in Figures \ref{fig:kouho_rekkyo} 
and \ref{fig:kouho_hantei}. 
Surface expressions, semantic constraints, 
referential properties, etc. are written as conditions 
in the {\sl Condition} part. 
Possible antecedents are written in the {\sl Possible-Antecedent} part. 
{\sl Point}s means the plausibility of the possible antecedent. 

An estimation of the referent is performed 
using the total scores of possible antecedents given 
by {\sf Candidate enumerating rule}s and {\sf Candidate judging rule}s. 
First, the system applies 
all {\sf Candidate enumerating rule}s to the anaphor 
and enumerates candidate antecedents having points. 
Next, 
the system applies 
all {\sf Candidate judging rule}s to all the candidate antecedents 
and sums the scores of all the candidate antecedents. 
Consequently, 
the system judges the candidate antecedent 
having the best score to be the proper antecedent. 
If several candidate referents have the best score, 
the candidate referent selected first in order\footnote{
The order is based on order applying rules. 
}
is judged to be 
the correct antecedent. 

We made 50 {\sf Candidate enumerating rule}s and 
10 {\sf Candidate judging rule}s for analyzing demonstratives, 
4 {\sf Candidate enumerating rule}s and 
6 {\sf Candidate judging rule}s for analyzing personal pronouns, 
and 19 {\sf Candidate enumerating rule}s and 
4 {\sf Candidate judging rule}s for analyzing zero pronouns. 
Some of the rules are described in the following sections. 

\section{\normalsize \bf Heuristic Rules for Demonstratives}

\label{sec:sijisi_ana}

We made heuristic rules for demonstratives 
by consulting the papers 
\cite{seiho1} \cite{hyasi2} \cite{sijisi_nihongogaku} \cite{sijisi} 
and by examining Japanese sentences by hand. 
Demonstratives 
have three categories: 
demonstrative pronouns,  
demonstrative adjectives,  and 
demonstrative adverbs. 
In the following sections, we explain the rules for 
analyzing demonstratives. 

\subsection{\small \bf Rule for Demonstrative Pronouns}
\label{sec:meishi_siji}

\subsection*{\small \bf \underline{Rule in the case when
the referent is a noun} \underline{phrase}}


\noindent
{\bf {\sf Candidate enumerating rule 1}}
\begin{indention}{0.8cm}\noindent
  When a pronoun is a demonstrative pronoun or 
  ``{\it sono} (of it) / {\it kono} (of this) / {\it ano} (of that)'',\\
  \{(A topic which has weight $W$ and distance $D$, \, $W-D-2$)\\
  (A focus which has weight $W$ and distance $D$, \, $W-D+4$)\}\\
  This bracketed expression represents 
  the lists of proposals in Figure \ref{fig:kouho_rekkyo}. 
  The definition and weight $W$ of 
  the topic and focus are shown in 
  Tables \ref{tab:5c_topic} and \ref{tab:5c_focus}. 
  The distance ($D$) is the number of topics and foci 
  between the demonstrative and the possible referent. 
  Since a demonstrative more often refers to 
  foci than a zero pronoun does, 
  we add the coefficient $-2$ or $+4$ as compared with 
  the heuristic rules in zero pronoun resolution. 
\end{indention}
\vspace{0.5cm}

The score (in other words, 
the certification value) of a candidate referent
depends on 
the weight of topics/foci and the physical distance 
between the demonstrative and the candidate referent. 

\subsection*{\small \bf \underline{Rule when 
the referent is a verb phrase}}


\noindent
{\bf {\sf Candidate enumerating rule 2}}
\begin{indention}{0.8cm}\noindent
  When a pronoun is  ``{\it kore}/{\it sore}/{\it are''} or 
  a demonstrative adjective,\\
  \label{kore_mae}
  \{(
  The previous sentence (or the verb phrase 
  which is a conditional form containing a conjunctive particle such as
  ``{\it ga} (but)'', `` {\it daga} (but)'', and ``{\it keredo} (but)''
  if the verb phrase is in the same sentence), 
   \,$15$)\}
\end{indention}
\vspace{0.5cm}

The following is an example 
of a pronoun referring to the verb phrase in the previous sentence. 
\begin{equation}
  \begin{minipage}[h]{14cm}
\footnotesize
  \begin{tabular}[t]{l@{ }l@{ }l@{ }l@{ }l@{ }l}
\footnotesize {\it tengu-wa} & \footnotesize {\it maenoban-noyouni} & \footnotesize {\it utattari} & \footnotesize {\it odottari} & \footnotesize {\it shihajimeta}.\\
\footnotesize (tengu) & \footnotesize (the previous night) & \footnotesize (sing) & \footnotesize (dance) & \footnotesize (begin to do) \\
\multicolumn{5}{p{7.5cm}}{\footnotesize 
(Tengus began singing  
and dancing just as they had done the previous night.)}\\
\end{tabular}

\vspace{0.1cm}

  \begin{tabular}[t]{l@{ }l@{ }l@{ }l@{ }l}
\footnotesize {\it ojiisan-wa} & \footnotesize \underline{{\it sore}}-{\it wo} & \footnotesize {\it mite}, & \footnotesize {\it kon'nahuuni} & \footnotesize {\it utai-hajimeta}.\\
 (the old man) & (it) & (see) & (as follows) & (begin to sing)\\
\multicolumn{5}{l}{\footnotesize 
(When the old man saw \underline{this}, he began to sing as follows.)}\\
\end{tabular}
  \end{minipage}
\label{eqn:sore_mite_utau}
\end{equation}
In these sentences, 
a demonstrative pronoun ``{\it sore} (it)'' refers to 
the event 
``{\it tengutachi-ga} {\it utattari} {\it odottari} {\it shi-hajimemashita}
(tengus began singing  
and dancing just as they had done the previous night.)''\footnote{
A tengu is a kind of monster.}.

\begin{table}[t]
\vspace*{-1cm}
  \footnotesize
  \leavevmode
  \begin{center}
    \caption{Points given in the case of demonstrative pronouns}
    \label{tab:hininshoudaimeisi_ruijido}
\begin{tabular}[c]{|l|@{\hspace{0.12cm}}r@{\hspace{0.12cm}}|@{\hspace{0.12cm}}r@{\hspace{0.12cm}}|@{\hspace{0.12cm}}r@{\hspace{0.12cm}}|@{\hspace{0.12cm}}r@{\hspace{0.12cm}}|@{\hspace{0.12cm}}r@{\hspace{0.12cm}}|@{\hspace{0.12cm}}r@{\hspace{0.12cm}}|@{\hspace{0.12cm}}r@{\hspace{0.12cm}}|@{\hspace{0.12cm}}r@{\hspace{0.12cm}}|}\hline
Sim.     & 0 & 1 & 2 & 3 & 4 & 5 & 6 & Exact\\\hline
Points   & 0 & 0 & $-$10 & $-$10 & $-$10 & $-$10& $-$10& $-$10\\\hline
\end{tabular}
\end{center}
\vspace{-0.3cm}
\hspace{0.5cm}Sim. = Simlarity level
\end{table}

\subsection*{\small \bf \underline{Rule using 
the feature that 
demonstrative} \underline{pronouns usually do not refer to people}}


\noindent
{\bf {\sf Candidate judging rule 1}}
\begin{indention}{0.8cm}\noindent
  When a pronoun is a demonstrative pronoun and 
  a candidate referent has a semantic marker {\sf HUM} (human), 
  it is given $-10$. 
  We used the 
  Noun Semantic Marker Dictionary \cite{imiso-in-BGH} 
  as a semantic marker dictionary\footnote{This 
    dictionary includes semantic categories shown in Table \ref{tab:bunrui_code_change}.}. 
\end{indention}

\begin{table}[t]
\vspace*{-0.5cm}
  \footnotesize
  \leavevmode
  \begin{center}
    \caption{Modification of category number of ``{\it bunrui} {\it goi} {\it hyou''}}
    \label{tab:bunrui_code_change}
\begin{tabular}[c]{|@{\hspace{0.15cm}}l@{\hspace{0.15cm}}|@{\hspace{0.15cm}}l@{\hspace{0.15cm}}|@{\hspace{0.15cm}}l@{\hspace{0.15cm}}|}\hline
Semantic marker   & Original       & Modified\\
                  &   code            &     code\\\hline
{\sf ANI}(animal)         &  156                & 511\\[0cm]
{\sf HUM}(human)         &  12[0-4]            & 52[0-4]\\[0cm]
{\sf ORG}(organization)   &  12[5-8]   & 53[5-8]\\[0cm]
{\sf PLA}(plant)         &  155                & 611\\[0cm]
{\sf PAR}(part of living thing)   &  157                & 621\\[0cm]
{\sf NAT}(natural)       &  152                & 631\\[0cm]
{\sf PRO}(products) &  14[0-9]            & 64[0-9]\\[0cm]
{\sf LOC}(location)   &  117,125,126        & 651,652,653\\[0cm]
{\sf PHE}(phenomenon)     &  150,151            & 711,712\\[0cm]
{\sf ACT}(action)   &  13[3-8]            & 81[3-8]\\[0cm]
{\sf MEN}(mental)         &  130                & 821\\[0cm]
{\sf CHA}(character)         &  11[2-58],158       & 83[2-58],839\\[0cm]
{\sf REL}(relation)         &  111                & 841\\[0cm]
{\sf LIN}(linguistic products)     &  131,132            & 851,852\\[0cm]
Others            &  110                & 861\\[0cm]
{\sf TIM}(time)         &  116                & a11\\[0cm]
{\sf QUA}(quantity)         &  119                & b11\\[0cm]\hline
\end{tabular}

``125'' and ``126'' are given two category numbers. 
\end{center}
\end{table}

\vspace{0.5cm}
\noindent
{\bf {\sf Candidate judging rule 2}}
\begin{indention}{0.8cm}\noindent
  When a pronoun is a demonstrative pronoun, 
  a candidate referent is given the points 
  in Table \ref{tab:hininshoudaimeisi_ruijido} 
  by using the highest semantic similarity 
  between the candidate referent and 
  the codes 
  \{5200003010 5201002060 5202001020 5202006115 5241002150 5244002100\}
  in ``Bunrui Goi Hyou (BGH)'' \cite{BGH}\footnote{
In BGH, each word has a number called {\it a category number}. 
In an electrical version of BGH, 
each word has a 10-digit category number. 
This 10-digit category number indicates 
seven levels of an {\sf is-a} hierarchy. 
The top five levels are expressed by 
the first five digits of a category number. 
The sixth level is expressed 
by the following two digits of a category number. 
The last level is expressed by the last three digits of a category number. }
 which signify human beings. 
  When we calculate the semantic similarity, 
  we use the modified code table 
  in Table \ref{tab:bunrui_code_change}.
  The reason for this modification is that 
  some codes in BGH \cite{BGH} are not suitable for semantic constraints. 
\end{indention}
\vspace{0.5cm}

These rules use the feature that 
a demonstrative pronoun rarely refers to people. 
This reduces the number of candidates of the referent. 
For example, we find 
``{\it sore} (it)'' in the following sentences refers to 
``{\it konpyuuta} (computer)'', 
because ``{\it sore} (it)'' can only refer to only a thing which is not human 
and the only noun which is near ``{\it sore} (it)'' 
and which is not human is ``{\it konpyuuta} (computer)''. 
\begin{equation}
  \begin{minipage}[h]{11.5cm}
\footnotesize
  \begin{tabular}[t]{llll}
{\it taroo-wa} & {\it saishin-no} & {\it konpyuuta-wo} & {\it kaimashita}.\\
 (Taroo) & (new) & (computer) & (buy)\\
\multicolumn{4}{l}{
(Taroo bought a new computer.)}\\
\end{tabular}

\vspace{0.1cm}

  \begin{tabular}[t]{llll}
{\it jon-ni} & {\it sassoku} & \underline{{\it sore}}-{\it wo} & {\it misemashita}.\\
 (John) & (at once) & (it) & (show)\\
\multicolumn{4}{l}{
([He] showed \underline{it} at once to John.)}
\end{tabular}
  \end{minipage}
\label{eqn:5c_sore_new_computer}
\end{equation}

\subsection*{\small \bf \underline{Rule with feature that 
``{\it koko''} and ``{\it soko''}} \underline{often refer to locations}}


\noindent
{\bf {\sf Candidate judging rule 3}}
\begin{indention}{0.8cm}\noindent
  When a pronoun is ``{\it koko} (here) / {\it soko} (there) / {\it asoko} (over there)'' 
  and a candidate referent has 
  a semantic marker {\sf LOC} (location), 
  the candidate referent is given $10$ points. 
\end{indention}

\begin{table}[t]
\vspace*{-0.5cm}
\footnotesize
  \leavevmode
  \begin{center}
    \caption{Points given demonstrative pronouns 
      which refer to places}
    \label{tab:bashomeisi_ruijido}
\begin{tabular}[c]{|l|r|r|r|r|r|r|r|r|}\hline
Sim. & 0 & 1 & 2 & 3 & 4 & 5 & 6 & Exact\\\hline
Points   & $-$10 & $-$5 & 0 & 5 & 10 & 10& 10& 10\\\hline
\end{tabular}
\end{center}
\end{table}

\vspace{0.5cm}
\noindent
{\bf {\sf Candidate judging rule 4}}
\begin{indention}{0.8cm}\noindent
  When a pronoun is ``{\it koko}/{\it soko}/{\it asoko''}, 
  a candidate referent is given the points in 
  Table \ref{tab:bashomeisi_ruijido} 
  based on the semantic similarity 
  between the candidate referent and 
  the codes 
  \{6563006010 6559005020 9113301090 9113302010 6471001030 6314020130\}
  which signify locations in BGH \cite{BGH}. 
\end{indention}
\vspace{0.5cm}

``{\it soko} (there)'' commonly refers to location. 
For example, 
``{\it soko''} in the following sentences refers to 
``{\it baiten} (shop)'' which signifies location. 
\begin{equation}
  \begin{minipage}[h]{11.5cm}
\footnotesize
  \begin{tabular}[t]{llll}
{\it koora-wo} & {\it kaini} & {\it baiten-ni} & {\it hairimashita}.\\
 (cola) & (buy) & (shop) & (enter)\\
\multicolumn{4}{l}{
(Taroo entered a shop to buy a cola.)}\\
\end{tabular}

\vspace{0.1cm}

  \begin{tabular}[t]{llll}
{\it jiroo-wa} & \underline{{\it soko-de}} & {\it guuzen} & {\it dekuwashimashita}.\\
 (Jiroo) & (there) & (by chance) & (meet)\\
\multicolumn{4}{l}{
(Jiroo met Taroo \underline{there} by chance.)}\\
\end{tabular}
  \end{minipage}
\label{eqn:soko_dekuwasu}
\end{equation}

\subsection*{\small \bf \underline{Rule when
``{\it kokode''} or ``{\it sokode''} is used as a} \underline{conjunction}}


\noindent
{\bf {\sf Candidate enumerating rule 3}}
\begin{indention}{0.8cm}\noindent
  When a pronoun is ``{\it kokode''} or ``{\it sokode''}, \\
  \{(the pronoun is used as a conjunction, \,$11$)\}
\end{indention}
\vspace{0.5cm}

This rule is for 
when ``{\it kokode} (here or then)'' or ``{\it sokode} (there or then)'' 
is used as a conjunction. 
If a word that signifies location 
is not found near ``{\it kokode''} or ``{\it sokode''}, 
the candidate listed by this rule has the highest score, and 
``{\it kokode''} or ``{\it sokode''} is judged to be a conjunction. 
By using this rule, 
``{\it sokode''}  in the following sentences is judged 
to be a conjunction. 
\begin{equation}
  \begin{minipage}[h]{11.5cm}
    \footnotesize
  \begin{tabular}[t]{lll}
{\it ojiisan-wa} & {\it tengu-ga} & {\it kowakunakunatte-imashita}.\\
(old man) & (tengu) & (lose all fear of)\\
\multicolumn{3}{l}{
(The old man lost all fear of the tengus.)}\\
\end{tabular}

\vspace{0.1cm}

  \begin{tabular}[t]{l@{ }l@{ }l@{ }l@{ }l}
\underline{{\it sokode}} & {\it ojiisan-wa} & {\it kakureteita} & {\it ana-kara} & {\it detekimashita}.\\
 (so) & (old man) & (be hiding) & (hole) & (leave)\\
\multicolumn{5}{l}{
(\underline{So}, he left the hole where he had been hiding.)}\\
\end{tabular}
  \end{minipage}
\label{eqn:soko_ojiisan_kakureru}
\end{equation}
This rule is necessary 
when the system translates ``{\it sokode''} into English, 
judges whether it is used as a demonstrative or as a conjunction, 
and translates it into ``there'' or ``then.'' 

\subsection*{\small \bf \underline{Rule when an anaphor does not have its} \underline{antecedent}}

\noindent
{\bf {\sf Candidate enumerating rule 5}}
\begin{indention}{0.8cm}\noindent
  When a pronoun is a demonstrative pronoun, 
  a demonstrative adverb, or a demonstrative adjective,\\
  \{(Introduce an individual, \,$10$)\}\\

This rule is used 
when there is no referent of a pronoun in the sentences. 
This rule makes the system 
introduce a certain individual. 
\end{indention}

\subsection{\small \bf Rule for Demonstrative Adjectives}
\label{sec:rentai}

Demonstrative pronouns 
such as ``{\it kono} (this)'', 
``{\it sono} (the)'', 
``{\it ano} (that)'', 
``{\it kon'na} (like this)'', 
and ``{\it son'na} (like it)'' 
are classified into two reference categories: 
{\it gentei}-reference and {\it daikou}-reference. 

In a {\it Gentei}-reference 
although a demonstrative adjective does not refer to an entity by itself, 
the phrase of ``demonstrative adjective $+$ noun phrase'' 
refers to the antecedent. 
For example 
``{\it kono} {\it ojiisan} (this old man)'' in the following sentences: 
\begin{equation}
  \begin{minipage}[h]{11.5cm}
    \footnotesize
  \begin{tabular}[t]{l@{ }l@{ }l@{ }l}
{\footnotesize {\it ojiisan-wa}} & {\footnotesize {\it tengutachi-no-maeni}} & {\footnotesize {\it deteitte}} & {\scriptsize {\it odori-hajimemashita}}\\
\footnotesize (old man) & \footnotesize (before the tengus) & \footnotesize (appear) & \footnotesize (begin to dance)\\
\multicolumn{4}{l}{
(He appeared before the tengus, 
and began to dance.)}\\
\end{tabular}

\vspace{0.1cm}

  \begin{tabular}[t]{l@{ }l@{ }l@{ }l@{ }l}
{\footnotesize {\it keredomo}} & {\footnotesize \underline{{\it kono} {\it ojiisan-wa}}} & {\footnotesize {\it uta-mo}} & {\footnotesize {\it odori-mo}} & {\footnotesize {\it hetakuso-deshita}}\\
 (but) & (this old man) & (sing) & (dance) & (poor)\\
\multicolumn{5}{p{7.5cm}}{\footnotesize
(But \underline{the old man} was a poor singer, 
and his dancing was no better.)}\\
\end{tabular}
  \end{minipage}
\label{eqn:kono_ojiisan_heta}
\end{equation}
In this example, 
although the demonstrative 
``{\it kono} (this)'' does not refer to ``{\it ojiisan} (old man)'' in the first sentence, 
the noun phrase 
``{\it kono} {\it ojiisan} (this old man)'' refers to ``{\it ojiisan} (old man)'' 
in the first sentence. 

{\it Daikou}-reference is 
a demonstrative adjective that refers to an entity. 
In this case, we can analyze 
``{{\it sono}} (the)'' as well as ``{{\it sore-no}} (of it)''. 
In the following sentences, 
``{{\it sono}}'' refers to ``{{\it tengu}}'' (tengus). 
It is an example of {\it daikou}-reference. 
\begin{equation}
  \begin{minipage}[h]{11.5cm}
    \footnotesize
  \begin{tabular}[t]{l@{ }l@{ }l@{ }l@{ }l}
{\footnotesize {\it mata}} & {\footnotesize {\it karasu-no-youna}} & {\footnotesize {\it kao-wo-shita}} & {\footnotesize {\it tengu-mo}} & {\footnotesize {\it imashita}}\\
 (also) & (like crows) & (with face) & (tengu) & (exist)\\
\multicolumn{5}{l}{\footnotesize
(There were also some tengus with faces like those of crows.)}\\
\end{tabular}

\vspace{0.1cm}

  \begin{tabular}[t]{l@{ }l@{ }l}
{\footnotesize \underline{{\it sono} {\it kuchi}}-{\it wa}} & {\footnotesize {\it torino-kuchibashi-noyouni}} & {\footnotesize {\it togatte-imashita}}\\
 (their mouths) & (like the beaks of birds) & (be pointed) \\
\multicolumn{3}{l}{
(\underline{Their mouths} were pointed like the beaks of birds.)}\\
\end{tabular}
  \end{minipage}
\label{eqn:sono_kuti}
\end{equation}

Rules for {\it gentei}-reference and {\it daikou}-reference are as follows:  

\vspace{0.5cm}
\noindent
{\bf {\sf Candidate enumerating rule 7}}
\begin{indention}{0.8cm}\noindent
  When a pronoun is ``demonstrative adjective + noun $\alpha$,''\\
  \{
  (the noun phrase containing a noun $\alpha$, \,$45$)\\
  (the topic which is a subordinate of noun $\alpha$ 
  and which has weight $W$ and distance $D$, \,$W-D+30$)\\
  (the focus which is a subordinate of noun $\alpha$ 
  and which has weight $W$ and distance $D$, \,$W-D+30$)\}
\end{indention}
\vspace{0.5cm}

  The relationships between a super-ordinate word and a subordinate word 
  are detected by judging 
  the last word in the definition of the word $\alpha$ in EDR Japanese word dictionary \cite{edr_tango_2.1} 
  to be the super-ordinate of the word $\alpha$. 
  
Because of this rule, 
when 
a pronoun is ``demonstrative adjective $+$ noun phrase $\alpha$'' 
and there is the same noun phrase $\alpha$ near it, 
it is judged to be ``{\it gentei}-reference'' 
and is selected as a candidate of the referent. 
When there is a subordinate of a noun phrase $\alpha$ near it, 
it is also selected as a candidate of the referent. 
These rules give higher points 
to a candidate referent than other rules do. 
The following is an 
example of 
the ``demonstrative adjective $+$ noun phrase $\alpha$'' 
referring to the subordinate of noun phrase $\alpha$. 
\begin{equation}
  \begin{minipage}[h]{11.5cm}
\footnotesize
  \begin{tabular}[t]{l@{ }l@{ }l@{ }l@{ }l}
{\footnotesize \it ojiisan-wa} & {\footnotesize \it toonoiteiku} & {\footnotesize \it tsuru-no} & {\footnotesize \it sugata-wo} & {\footnotesize \it miokurimashita}.\\
 (old man) & (recede) & (crane) & (figure) & (watch)\\
\multicolumn{5}{l}{\footnotesize
(The old man watched the receding figure of the crane.)}\\
\end{tabular}

\vspace{0.1cm}

  \begin{tabular}[t]{llll}
``\underline{{\it ano} {\it tori}}-{\it wo} & {\it tasukete} & {\it yokatta''} {\it to} & {\it iimashita}.\\
 (that bird) & (save) & (glad) & (say)\\
\multicolumn{4}{l}{\footnotesize
(``I'm glad I saved \underline{that bird},'' said the old man to himself.)}\\
\end{tabular}
  \end{minipage}
\label{eqn:ano_tori}
\end{equation}
In this example, 
the underlined ``{{\it ano} {\it tori}} (that bird)'' refers to 
a subordinate ``{{\it tsuru}} (crane)'' in the previous sentence. 

\begin{table}[t]
\vspace*{-0.5cm}
\footnotesize
  \leavevmode
  \begin{center}
    \caption{Points given to so-series demonstrative adjective}
    \label{tab:sokei_meishi_anob_ruijido}
\begin{tabular}[c]{|l|r|r|r|r|r|r|r|r|}\hline
Sim. & 0   & 1  & 2  & 3 & 4 & 5 & 6 & Exact \\\hline
Points   & $-$10 & $-$2 & $-$1 & 0 & 1 & 2 & 3 & 4\\\hline
\end{tabular}
\end{center}
\end{table}

\subsection*{\small \bf \underline{Rules for 
{\it daikou}-reference of so-series} \underline{demonstrative adjective}}


\noindent
{\bf {\sf Candidate judging rule 5}}
\begin{indention}{0.8cm}\noindent
  When a pronoun is a {\it so}-series demonstrative adjective, 
  the system consults examples of 
  the form ``noun X {{\it no}} noun Y'' 
  whose noun Y is modified by the pronoun, 
  and gives a candidate referent 
  the points 
  in Table~\ref{tab:sokei_meishi_anob_ruijido}
  according to the similarity between 
  the candidate referent and noun X 
  in ``Bunrui Goi Hyou'' \cite{BGH}. 
  The Japanese Co-occurrence Dictionary \cite{edr_kyouki_2.1} is used 
  as a source of examples of ``X {\it no} Y''. 
\end{indention}
\vspace{0.5cm}

This rule is for checking 
the semantic constraint 
(For a {\it daikou}-reference, candidates of the referent 
are selected by {\sf Candidate enumerating rule 1} in Section \ref{sec:meishi_siji}.).

We explain how to use the rule 
in the underlined ``{{\it sono}} (the)'' 
in the sentences (\ref{eqn:sono_kuti}). 
First, the system gathers 
examples of the form ``Noun X {{\it no} {\it kuchi}} (mouth of Noun X)''. 
Table \ref{fig:meishi_A_kuti} shows 
some examples of ``Noun X {{\it no} {\it kuchi}} (mouth of Noun X)'' in the Japanese Co-occurrence Dictionary \cite{edr_kyouki_2.1}. 
Next, the system checks the semantic similarity 
between candidate referents and Noun X, and 
judges the candidate referent having a higher similarity 
to be a better candidate referent. 
In this example, 
``{{\it tengu}}'' is semantically similar to Noun X 
in that they are both living things. 
Finally, the system selects 
``{{\it tengu}}'' as the proper referent. 

\begin{table}[t]
\vspace*{-0.5cm}
\footnotesize
\begin{center}
  \caption{Examples of the form ``the mouth of Noun X'' }
  \label{fig:meishi_A_kuti}
\begin{tabular}[c]{|p{7.5cm}|}\hline
Examples of Noun X \\\hline
{{\it hukuro}} (sack),  {{\it ruporait\=a}} (documentary writer) 
{{\it iin}} (member), {{\it akachan}} (baby), {{\it kare}} (he)\\\hline
\end{tabular}
\end{center}
\end{table}

\begin{table}[t]
\vspace*{-0.5cm}
\footnotesize
  \leavevmode
  \begin{center}
  \caption{Points given in the case of non-{\it so}-series demonstrative adjective}
    \label{tab:akei_meishi_anob_ruijido}
\begin{tabular}[c]{|l|@{\hspace{0.12cm}}r@{\hspace{0.12cm}}|@{\hspace{0.12cm}}r@{\hspace{0.12cm}}|@{\hspace{0.12cm}}r@{\hspace{0.12cm}}|@{\hspace{0.12cm}}r@{\hspace{0.12cm}}|@{\hspace{0.12cm}}r@{\hspace{0.12cm}}|@{\hspace{0.12cm}}r@{\hspace{0.12cm}}|@{\hspace{0.12cm}}r@{\hspace{0.12cm}}|@{\hspace{0.12cm}}r|}\hline

Sim. & 0   & 1   & 2   & 3   & 4   & 5 & 6 & Exact\\\hline
Points   & $-$30 & $-$30 & $-$30 & $-$30 & $-$10 & $-$5& $-$2& 0\\\hline
\end{tabular}
\end{center}
\end{table}

\subsection*{\small \bf \underline{Rules when non-{\it so}-series demonstrative has} 
\underline{{\it daikou}-reference}}


\noindent
{\bf {\sf Candidate judging rule 6}}
\begin{indention}{0.8cm}\noindent
  When a pronoun is a non-{\it so}-series demonstrative adjective, 
  the system consults examples of 
  the form ``Noun X {{\it no} (of)} Noun Y (Y of X)'' 
  whose Noun Y is modified by the pronoun, 
  and gives candidate referents 
  the points 
  in Table \ref{tab:akei_meishi_anob_ruijido}
  according to the similarity between 
  the candidate referent and noun X 
  in ``Bunrui Goi Hyou'' \cite{BGH}. 
  Since a non-{\it so}-series demonstrative adjective 
  rarely is a {\it daikou} reference 
  \cite{seiho1}
  \cite{yamamura92_ieice}, 
  the number of points is footnotesizeer than in the case of the {\it so}-series. 
\end{indention}

\subsection*{\small \bf \underline{Rule when a pronoun refers to a verb phrase}}


Like a demonstrative pronoun, 
a demonstrative adjective can refer to the meaning 
of the verb phrase in the previous sentence. 
This case is resolved by {\sf Candidate enumerating rule 2} 
in Section \ref{sec:meishi_siji}. 

\subsection*{\small \bf \underline{Rule for ``{\it kon'na} noun'' (noun like this)}}



``{\it kon'na} noun'' can also refer to the next sentences 
in addition to a noun phrase and the previous sentences. 
\begin{equation}
  \begin{minipage}[h]{11.5cm}
\footnotesize
  \begin{tabular}[t]{l@{ }l@{ }l@{ }l}
{\it ojiisan-wa} & {\it odorinagara} & \underline{{\it kon'na} {\it uta}}-{\it wo} & {\it utaimashita}.\\
 (old man) & (dance) & (song like this) & (sing)\\
\multicolumn{4}{l}{
(As he danced, he sang \underline{the following song}: )}\\
\end{tabular}

\vspace{0.1cm}

  \begin{tabular}[t]{llll}
``{\it tengu} & {\it tengu} & {\it hachi} {\it tengu}.\\
 (tengu) & (tengu) & (eight tengu)\\
\multicolumn{4}{l}{
(```Tengu,' `tengu,' eight `tengus.''')}\\
\end{tabular}
  \end{minipage}
\label{eqn:konana_kouhou}
\end{equation}
In the above example, 
``{\it kon'na} {\it uta} (song like this)'' refers to the next sentence 
``{\it tengu}, {\it tengu}, {\it hachi} {\it tengu}.''

\begin{table}[t]
\vspace*{-0.5cm}
\footnotesize
\begin{center}
  \caption{Results of investigating 
whether ``{\it kon'na} noun'' (noun like this) refers to the previous 
or next sentences}
  \label{fig:konna_meishi_joshi_tyosha}
\begin{tabular}[c]{|l|r|r|}\hline
  \multicolumn{1}{|l|}{Postpositional particle}  & \multicolumn{1}{|l|}{previous} & \multicolumn{1}{|l|}{next}\\[-0.1cm]
  \multicolumn{1}{|l|}{}  & \multicolumn{1}{|l|}{sentence} & \multicolumn{1}{|l|}{sentence}\\\hline
{\it wa} (topic) & 9 & 0\\\hline
{\it wa-nai}  & 5 & 0\\\hline
{\it ni} (indirect object) & 17 & 0\\\hline
{\it ni-mo} & 1 & 0\\\hline
{\it ni-wa} & 2 & 0\\\hline
{\it de} (place)   & 15 & 0\\\hline
{\it de-wa} & 5 & 0\\\hline
{\it no} (possessive)   & 9 & 0\\\hline
{\it sura}   & 2 & 0\\\hline
{\it ga} (subject)   & 27 & 22\\\hline
{\it wo} (object)   & 43 & 26\\\hline
{\it mo} (also)  & 2 & 4\\\hline
{\it de-wa-nai} & 0 &1\\\hline
Total & 137& 53\\\hline
\end{tabular}
\end{center}
\end{table}

But 
we cannot decide whether ``{\it kon'na} $+$ noun'' (noun like this) 
refers to the previous or next sentences 
only by the expression of 
``{\it kon'na} $+$ noun'' (noun like this) itself. 
To make the decision, 
we gathered 317 sentences containing ``{\it kon'na}'' (like this) 
from about 60,000 sentences in 
Japanese essays and editorials, 
and counted the total frequency of cases in which 
``{\it kon'na''}  refers to the previous and 
next sentences. 
The results are shown in Table \ref{fig:konna_meishi_joshi_tyosha}. 
This table indicates that 
``{\it kon'na} $+$ noun'' followed by other particles, 
specifically ``{{\it ga}}'' and ``{{\it wo}},'' 
which are used when representing new information, 
very often refers to the previous sentence. 
Therefore, the system judges that 
the desired antecedent is the previous sentence. 
When ``{\it kon'na} noun'' is followed by the particles 
``{{\it ga}}'' or ``{{\it wo}},'' 
the proper referent is determined 
by the expression in quotation marks (``,''). 

\subsection{\small \bf Rule for Demonstrative Adverbs}

\subsection*{\small \bf \underline{Rule when 
{\it so}-series demonstrative adverb}\\ 
\underline{refers to the previous sentences}}


\noindent
{\bf {\sf Candidate enumerating rule 9}}
\begin{indention}{0.8cm}\noindent
  When an anaphor is 
  a {\it so}-series demonstrative adverb such as ``{{\it sou} (so)},''\\ 
  \{(the previous sentences, \,$30$)\}
\end{indention}

The following is an example. 
\begin{equation}
  \begin{minipage}[h]{11.5cm}
\footnotesize
  \begin{tabular}[t]{llll}
``{\footnotesize {\it tengu}} & {\footnotesize {\it tengu}} & {\footnotesize {\it hachi} {\it tengu}}.''\\
 (tengu) & (tengu) & (eight tengu)\\
\multicolumn{4}{l}{
(```Tengu,' `tengu,' eight `tengus.''')}\\
\end{tabular}

\vspace{0.1cm}

  \begin{tabular}[t]{l@{ }l@{ }l@{ }l@{ }l}
{\footnotesize \underline{{\it sou}} {\it utatta-nowa}} & {\footnotesize {\it sokoni}} & {\footnotesize {\it hachihiki-no}} & {\footnotesize {\it tengu-ga}} & {\footnotesize {\it itakara-desu}}.\\
 (sing so) & (there) & (eight) & (tengu) & (exist)\\
\multicolumn{5}{l}{
(He sang \underline{so} because he counted eight of them there. )}\\
\end{tabular}
  \end{minipage}
\label{eqn:sono_utau}
\end{equation}
``{{\it sou}} (so)'' refers to the previous sentence 
``{\it tengu} {\it tengu} {\it hachi} {\it tengu''}.

\subsection*{\small \bf \underline{Rule when 
{\it so}-series demonstrative adverb}\\ 
\underline{cataphorically Refers to 
the Verb Phrase} \underline{in the Same Sentence}}

\noindent
{\bf {\sf Candidate enumerating rule 10}}
\begin{indention}{0.8cm}\noindent
  When an anaphor is ``{\it sou}/{\it soushite}/{\it sonoyouni''} 
  and
  is in the subordinate clause 
  which has a conjunctive particle such as
  ``{\it ga''}, ``{\it daga''}, and ``{\it keredo''} 
  or an adjective conjunction such as ``{\it youni''},\\
  \{(the main clause, \,$45$)\}\\
\end{indention}

\section{\normalsize \bf Heuristic Rule for Personal Pronouns}
\label{sec:pro_ana}

\noindent
{\bf {\sf Candidate enumerating rule 1}}
\begin{indention}{0.8cm}\noindent
  When an anaphor is a first personal pronoun, \\
  \{(the first person (the speaker) in the context, \,$25$)\}
\end{indention}

\vspace{0.5cm}
\noindent
{\bf {\sf Candidate enumerating rule 2}}
\begin{indention}{0.8cm}\noindent
  When an anaphor is a second personal pronoun, \\
  \{(the second person (the hearer) in the context, \,$25$)\}
\end{indention}
\vspace{0.5cm}


A first or second personal pronoun is often presented 
in quotations, 
and can be resolved 
by estimating 
the first person (speaker) or the second person (hearer) 
in advance. 
The estimation of 
the first person and the second person is performed 
by regarding the {\it ga}-case ({\sf subjective}) 
and {\it ni}-case ({\sf objective}) components 
of the verb phase representing the speaking action of the quotation 
as the first and second persons, respectively. 
The detection of the verb phase representing the speaking action 
is performed as follows. 
If the quotation is followed by a speaking action verb phrase 
such as ``{{\it to} {\it itta}} (was said),'' 
the verb phrase is regarded as the verb phase 
representing the speaking action. 
Otherwise, the last verb phrase in the previous sentence is 
regarded as the verb phase 
representing the speaking action. 
For example, 
the second personal pronoun ``{{\it omaesan}} (you)'' in the following sentences 
refers to the second person ``{{\it ojiisan}} (the old man)'' 
in this quotation. 
\begin{equation}
  \begin{minipage}[h]{11.5cm}
    \footnotesize
  \begin{tabular}[t]{lll}
``{\it asu}, & {\it mata} & {\it mairimasuyo}.'' {\it to},\\
 (tomorrow) & (again) & (come) \\
\multicolumn{3}{l}{
(``I'll come again tomorrow,'')}\\
\end{tabular}

\vspace{0.1cm}

  \begin{tabular}[t]{lll}
{\it ojiisan-wa} & {\it yakusoku-shimashita}.\\
 (old man) & (promise) \\
\multicolumn{3}{l}{
(promised the old man.)}\\
\end{tabular}

\vspace{0.1cm}

  \begin{tabular}[t]{lll}
``{\it mochiron} & \underline{{\it omaesan}}-{\it wo} & {\it utagauwakedewanainodaga},''\\
 (of course) & (you) & (don't mean to doubt) \\
\multicolumn{3}{l}{
(``Of course, we don't mean to doubt \underline{you},'')}\\
\end{tabular}

\vspace{0.1cm}

  \begin{tabular}[t]{lll}
{\it tengu-ga} & \underline{{\it ojiisan-ni}} & {\it iimashita}.\\
 (tengu) & (old man) & (said) \\
\multicolumn{3}{l}{
(said one of the ``tengu'' to \underline{the old man}.)}\\
\end{tabular}
  \end{minipage}
\label{eqn:ojiisan_mairu_omae}
\end{equation}
The second person in the quotation is estimated to be ``{{\it ojiisan}}'' 
because the {\it ni}-case component of 
the verb phrase ``{{\it iimashita}} (said)'' 
representing the speaking action of the quotation is 
``{{\it ojiisan}}''. 

\begin{figure}[t]
\vspace*{-0.5cm}
  \footnotesize
  \leavevmode
  \begin{center}
\fbox{
    \begin{minipage}[c]{7.5cm}
    \begin{tabular}[t]{lll}
\underline{{\it ojiisan}}-{\it wa} & {\it jimen-ni} & {\it koshi-wo-oroshimashita}.\\
(old man) & (ground) & (sit down)\\
\multicolumn{3}{l}{
(\underline{The old man} sat down on the ground.)}\\[0.1cm]
\end{tabular}
\begin{tabular}[t]{lll}
{\it yagate} & (\underline{{\it ojiisan}}-{\it wa}) & {\it nemutte-shimaimashita}.\\
(soon) & (old man) & (fall asleep)\\
\multicolumn{3}{l}{
(\underline{He} soon fell asleep.)}\\
\end{tabular}\\[0.1cm]

\hspace*{.5cm}
\begin{tabular}[c]{ll}
\multicolumn{2}{l}{\hspace*{-.5cm}{\bf Semantic Marker}}\\[0.1cm]
{\sf HUM}/{\sf ANI} {\it ga} ({\sf agent}) &{\it nemuru} (sleep)\\[0.1cm]
\multicolumn{2}{l}{\hspace*{-.5cm}{\bf Example}}\\[0.1cm]
{\it kare} (he)/ {\it inu} (dog) {\it ga} ({\sf agent}) &{\it nemuru} (sleep)\\
\end{tabular}

\vspace{-0.2cm}

\caption{How to check semantic constraint}
\label{fig:datousei_hantei_rei}
\end{minipage}
}
  \end{center}
\end{figure}

\vspace{0.5cm}
\noindent
{\bf {\sf Candidate enumerating rule 3}}
\begin{indention}{0.8cm}\noindent
  When an anaphor is a third personal pronoun, \\
  \{(a first person, \,$-10$) (a second person, \,$-10$)\}
\end{indention}
\vspace{0.5cm}

\section{\normalsize \bf Heuristic Rule for Zero Pronoun}
\label{sec:zero_ana}

\subsection*{\small \bf \underline{Rule proposing candidate referents 
of general} \underline{zero pronoun}}


\noindent
{\bf {\sf Candidate enumerating rule 1}}
\begin{indention}{0.8cm}\noindent
  When a zero pronoun is a {\it ga}-case component, \\
  \{(A topic which has weight $W$ and distance $D$, \,$W-D*2+1$)\\
  (A focus which has weight $W$ and distance $D$, \,$W-D+1$)\\
  (A subject of a clause coordinately connected to the clause containing the anaphor, \,25)\\
  (A subject of a clause subordinately connected to the clause containing the anaphor, \,23)\\
  (A subject of a main clause whose 
  embedded clause contains the anaphor, \,22)\}
\end{indention}

\vspace{0.5cm}
\noindent
{\bf {\sf Candidate enumerating rule 2}}
\begin{indention}{0.8cm}\noindent
  When a zero pronoun is not a {\it ga}-case component, \\
  \{(A topic which has weight $W$ and distance $D$, \,$W-D*2-3$)\\
  (A focus which has weight $W$ and distance $D$, \,$W-D*2+1$)\}
\end{indention}
\vspace{0.5cm}

\subsection*{\small \bf \underline{Rule using semantic relation 
to verb phrase}}


\noindent
{\bf {\sf Candidate judging rule 1}}
\begin{indention}{0.8cm}\noindent
  When a candidate referent of a case component (a zero pronoun) 
  does not satisfy the semantic marker of the case component 
  in the case frame, 
  it is given $-5$. 
\end{indention}

\vspace{0.5cm}
\noindent
{\bf {\sf Candidate judging rule 2}}
\begin{indention}{0.8cm}\noindent
  A candidate referent of a case component (a zero pronoun) is 
  given the points in Table \ref{tab:yourei_ruijido}
  by using the highest semantic similarity 
  between the candidate referent and examples 
  of the case component in the case frame. 
\end{indention}
\vspace{0.5cm}

\begin{table}[t]
\vspace*{-0.5cm}
\footnotesize
  \leavevmode
  \begin{center}
    \caption{Points given by a verb-noun relationship}
    \label{tab:yourei_ruijido}
\begin{tabular}[c]{|l|r|r|r|r|r|r|r|r|}\hline
Sim. & 0 & 1 & 2 & 3 & 4 & 5 & 6 & Exact\\\hline
Points   & $-$10 & $-$2 & 1 & 2 & 2.5& 3 & 3.5 & 4\\\hline
\end{tabular}
\end{center}
\end{table}

These two rules are 
for checking 
the semantic constraint 
between the candidate referent and 
the verb phrase which has the candidate referent in its case component. 
{\sf Candidate judging rule 1} 
checks semantic constraints by using semantic markers. 
{\sf Candidate judging rule 2} 
checks semantic constraints by using examples. 
Figure \ref{fig:datousei_hantei_rei} explains how to check 
semantic constraints in the example sentences. 

\begin{figure*}[t]
  \footnotesize
  \begin{center}
\fbox{
\begin{minipage}[h]{15cm}

\hspace*{-0.3cm}
\begin{tabular}[t]{l@{ }l@{ }l@{ }l}
{\it doru} {\it souba-wa} & {\it kitai-kara} & 130-{\it yen-dai-ni} & {\footnotesize {\it joushoushita}}.\\
(dollar)   & (the expectations) & (130 yen) & (surge)\\
\multicolumn{4}{l}{
(The dollar has since rebounded to
about 130 yen because of the expectations.)}\\
\end{tabular}

\vspace{0.1cm}

\hspace*{-0.3cm}
\begin{tabular}[t]{l@{ }l@{ }l@{ }l}
\underline{{\it kono} {\it doru-daka-wa}} & {\footnotesize {\it oushuu-tono}} & {\footnotesize {\it kankei-wo}} & {\footnotesize {\it gikushaku-saseteiru}.}\\
(the dollar's surge) & (Europe) & (relation) & (strain)\\
\multicolumn{4}{l}{
(\underline{The dollar's surge} is straining relations with Europe.)}\\
\end{tabular}

\vspace{0.1cm}

{\renewcommand{\arraystretch}{1.2}

\begin{tabular}[h]{|l|c|c|c|c|c|}\hline
Rule                   & \multicolumn{5}{c|}{Score of each candidate (points)}\\\hline
                       & { the previous} & new & 130 {\it yen} & {\it kitai} & { {\it dorusouba}}\\[-0.1cm]
                       & sentence &  { individual} & (130 yen) & ({expectations}) & (dollar)\\\hline
{ \sf Candidate enumerating rule 2}       & 15   &            &         &        &         \\[-0.1cm]
{ \sf Candidate enumerating rule 5}    &      &    10      &         &        &         \\[-0.1cm]
{ \sf Candidate enumerating rule 1}       &      &            &   17    &  15    &  15     \\[-0.1cm]
{ \sf Candidate judging rule 6}      &      &            &  $-30$  &  $-30$ &  $-30$  \\\hline
Total score             & 15   &    10      &  $-13$  &  $-15$ &  $-15$  \\\hline
\end{tabular}}

\vspace{-0.2cm}

\caption{Example of resolving demonstrative ``{\it kono} (this)''}
\label{tab:5c_dousarei}
\end{minipage}
}
\end{center}
\end{figure*}


\begin{table*}[t]
\vspace*{-0.5cm}
\footnotesize
\begin{center}
\fbox{
  \begin{minipage}[h]{16cm}
\vspace*{-0.3cm}
\footnotesize
  \begin{center}
    \caption{Results}
    \label{tab:5c_sougoukekka}
\begin{tabular}[c]{|l|r@{ }c|r@{ }c|r@{ }c|r@{ }c|}\hline
\multicolumn{1}{|p{2cm}|}{Text}&
\multicolumn{2}{c|}{demonstrative}&
\multicolumn{2}{c|}{personal pronoun}&
\multicolumn{2}{c|}{zero pronoun}&
\multicolumn{2}{c|}{total score}\\\hline
Training    &  87\% &  (41/47)  & 100\% &   (9/ 9)   &  86\% & (177/205) &  87\% & (227/261) \\\hline
Test  &  86\% &  (42/49)  &  82\% &   (9/11)   &  76\% & (159/208) &  78\% & (210/268) \\\hline
\end{tabular}
\end{center}
\vspace{-0.3cm}
{The points given in each rule 
are manually adjusted by using the training sentences. \\
Training sentences \{example sentences (43 sentences), a folk tale ``{\it kobutori} {\it jiisan''} \cite{kobu} (93 sentences), an essay in ``{\it tenseijingo''} (26 sentences), an editorial (26 sentences), an article in ``Scientific American (in Japanese)''(16 sentences)\}\\
Test sentences \{a folk tale ``{\it tsuru} {\it no} {\it ongaeshi''} \cite{kobu} (91 sentences), two essays in ``{\it tenseijingo''} (50 sentences), an editorial (30 sentences), articles in ``Scientific American (in Japanese)'' (13 sentences)\}
}
  \end{minipage}
}
\end{center}
\end{table*}

In the method using semantic markers, 
a candidate referent is the proper referent 
if one of the semantic markers belonging to the candidate referent 
is equal or subordinate to the semantic marker of the case component. 
For example, 
with respect to the zero pronoun in Figure \ref{fig:datousei_hantei_rei}, 
since the {\it ga}-case component 
in the verb ``{{\it nemuru}} (sleep)'' 
has the semantic markers {\sf HUM} (human being) and {\sf ANI} (animal) 
and since ``{\it ojiisan} (old man)'' has the semantic marker {\sf HUM}, 
the proper referent is judged to be ``{\it ojiisan.''} 

In the example-based method, 
the validity of a candidate referent 
is decided by 
the semantic similarity 
between the candidate referent and 
the examples of the case component in the verb case frame. 
The higher the semantic similarity is, 
the greater the validity is. 
For example, 
with respect to a zero pronoun in Figure \ref{fig:datousei_hantei_rei}, 
since the examples of the {\it ga}-case are 
``{{\it kare}} (he)'' and ``{{\it inu}} (dog),''  
and since ``{\it ojiisan} (old man)'' 
is semantically similar to ``{{\it kare}} (he)'', 
the proper referent is ``{\it ojiisan} (old man).'' 

These rules, which use semantic relationships to verbs,  
are also used in the estimation of the referent of 
demonstratives and  personal pronouns. 

\subsection*{\small \bf \underline{Rule using the feature that 
it is difficult for} \underline{a noun phrase 
to be filled in multiple case} 
\underline{components of the same verb}}


\noindent
{\bf {\sf Candidate enumerating rule 4}}
\begin{indention}{0.8cm}\noindent
  When there is ``Noun X'' in another case component 
  of the verb which has the analyzed case component 
  (the analyzed zero pronoun), 
  \{(Noun X, \,$-20$)\}
\end{indention}

\subsection*{\small \bf \underline{Rule using empathy}}

This rule is based on empathy theory \cite{kameyama1}. 
When an anaphor is a {\it ga}-case zero pronoun 
whose verb is followed by an auxiliary verb 
such as ``{{\it kureru}}'' or ``{{\it kudasaru}},'' 
the {\it ni}-case zero pronoun is analyzed first, 
and 
it is filled with 
the noun phrase that has high empathy such as the topic, 
and 
a {\it ga}-case zero pronoun is filled with 
another noun phrase.

\begin{table*}[t]
\vspace*{-0.5cm}
\footnotesize
  \begin{center}
\footnotesize
    \caption{Detailed results for demonstrative}
    \label{tab:sijisi_kekka}
\begin{tabular}[c]{|l|r@{ }c|r@{ }c|r@{ }c|r@{ }c|}\hline
\multicolumn{1}{|p{2cm}|}{Text}&
\multicolumn{2}{c|}{demonstrative }&
\multicolumn{2}{c|}{demonstrative }&
\multicolumn{2}{c|}{demonstrative }&
\multicolumn{2}{c|}{total score}\\[-0.1cm]
\multicolumn{1}{|p{2cm}|}{}&
\multicolumn{2}{c|}{pronoun}&
\multicolumn{2}{c|}{adjective}&
\multicolumn{2}{c|}{adverb}&
\multicolumn{2}{c|}{}\\\hline
Training   &  83\% &  (15/18)  &  86\% &   (19/22)   & 100\% & (7/7) &  87\% & (41/47) \\\hline
Test  &  82\% &  (14/17)  &  88\% &   (23/26)   &  83\% & (5/6) &  86\% & (42/49) \\\hline
\end{tabular}
\end{center}
\end{table*}

\section{\normalsize \bf Experiment and Discussion}
\label{sec:jikken}

\subsection{\small \bf Experiment}

Before pronoun resolution, 
sentences were transformed into a case structure 
by a case structure analyzer \cite{csan2_ieice}. 
The errors made by the structure analyzer were corrected by hand. 
We used IPAL dictionary \cite{ipal} as a verb case frame dictionary. 
We put together 
the case frames of the verb phrases which were not contained in this dictionary 
by consulting a large amount of linguistic data. 

An example of resolving the demonstrative ``{\it kono} (this)'' 
is shown in Figure \ref{tab:5c_dousarei}, 
which shows that 
the referent of the noun phrase ``{{\it kono} {\it dorudaka}} (this dollar's surge)'' 
was properly judged to be the previous sentence. 

\begin{table*}[t]
\vspace*{-0.8cm}
\footnotesize
  \begin{center}
    \caption{Results of comparison between semantic marker and example-base}
    \label{tab:yourei_taishou}
\begin{tabular}[c]{|l|r@{}c|r@{}c|r@{}c|r@{}c|r@{}c|}\hline
& \multicolumn{2}{|c|}{Method 1} & 
 \multicolumn{2}{c|}{Method 2} & 
 \multicolumn{2}{c|}{Method 3} & 
 \multicolumn{2}{c|}{Method 4} & 
 \multicolumn{2}{c|}{Method 5}\\\hline 
Demonstrative & 
87\% &  (41/47)  &  83\% &  (39/47)  &  87\% &  (41/47)  &  83\% &  (39/47)  &  79\% &  (37/47)  \\\cline{2-11}
& 86\% &  (42/49)  &  88\% &  (43/49)  &  88\% &  (43/49)  &  84\% &  (41/49)  &  86\% &  (42/49)  \\\hline
Personal pronoun & 
100\% &   (9/ 9)   & 100\% &   (9/ 9)  & 100\% &   (9/ 9)  & 100\% &   (9/ 9)  &  89\% &   (8/ 9)   \\\cline{2-11}
& 82\% &   (9/11)   &  64\% &   (7/11)  &  82\% &   (9/11)  &  55\% &   (6/11)  &  64\% &   (7/11)  \\\hline
Zero pronoun & 
 86\% & (177/205)  &  83\% & (171/205) &  86\% & (176/205) &  82\% & (169/205) &  66\% & (135/205) \\\cline{2-11}
& 76\% & (159/208)  &  76\% & (158/208) &  79\% & (164/208) &  75\% & (155/208) &  63\% & (131/208) \\\hline
\end{tabular}
  \end{center}
\vspace{-0.3cm}
{\footnotesize
\hspace*{1cm}
Method 1 : Using both semantic marker and example

\hspace*{1cm}
Method 2 : Using semantic marker

\hspace*{1cm}
Method 3 : Using example (using modified codes of {\it bunrui} {\it goi} {\it hyou})

\hspace*{1cm}
Method 4 : Using example (using original codes of {\it bunrui} {\it goi} {\it hyou})

\hspace*{1cm}
Method 5 : Using neither semantic marker nor example
}
\end{table*}

By {\sf Candidate enumerating rule 2} in Section \ref{sec:sijisi_ana}, 
the system took a candidate ``the previous sentence'' 
and gave it 15 points. 
By {\sf Candidate enumerating rule 5} in Section \ref{sec:sijisi_ana}, 
the system took a candidate 
``new individual'' and gave it 10 points. 
By {\sf Candidate enumerating rule}1 in Section \ref{sec:sijisi_ana}, 
the system took three candidates,  
``130 {\it yen} (130 yen)'', ``{\it kitai} (expectations)'', 
and ``{\it dorusouba} (dollar)'', 
and gave them 17, 15, and 15 points, respectively. 
The system applied {\sf Candidate judging rule 6} to them. 
This uses examples of ``X {\it no} Y''. 
In this case, 
it used examples of ``X {\it no} {\it dorudaka} (the dollar's surge of X)''. 
The only example noun phrase X of 
this form ``X {\it no} {\it dorudaka''} in the EDR occurrence dictionary 
was ``{\it saikin} (recently)''. 
All three candidates, 
``130 {\it yen} (130 yen)'', ``{\it kitai} (expectations)'', 
and ``{\it dorusouba} (dollar)'', 
were low in similarity to ``{\it saikin} (recently)'' in ``Bun Rui Goihyou'', 
and were given $-30$ points by Table \ref{tab:akei_meishi_anob_ruijido}. 
Two candidates, ``the previous sentence'' and 
``new individual'', 
so they are not noun phrases, 
and were not given points by {\sf Candidate judging rule 6}. 
As a result, 
``the previous sentence'' had the highest score and 
was judged to be the proper referent. 

We show 
the results of our resolution 
of demonstratives, personal pronouns, and zero pronouns 
in Table \ref{tab:5c_sougoukekka}. 
The detailed results for demonstratives 
are shown in Table \ref{tab:sijisi_kekka}. 
The precision rate of zero pronouns is in the case when 
the system knows whether the zero pronoun has a referent or not 
in advance. 

\subsection{\small \bf Discussion}

With respect to demonstratives, 
the precision rate was over 80\% even in the test sentences. 
This indicates that the rules used in this system are effective. 
But since Japanese demonstratives are classified into many kinds, 
the precision may be increased 
by making more detailed rules. 
In this work 
we used the feature that 
``{{\it kono}} (this)'' rarely functions as a {\it daikou}-reference. 
There were four cases analyzed correctly because of this rule. 

With respect to personal pronouns, 
since only first and second personal pronouns appeared  
in the texts used in the experiment, 
almost all of the personal pronouns were resolved correctly 
by estimating the first and second persons in the quotation. 
The main reason for the errors in the personal pronoun resolution 
is that 
the {\it ni}-case zero pronoun was resolved incorrectly 
and the second person was estimated incorrectly. 

There are several reasons for the errors of the zero pronoun resolution: 
there are errors in Japanese thesaurus 
``Bunrui goi hyou'', Noun Semantic Marker Dictionary, 
and Case Frame Dictionary. 

\subsection{\small \bf Comparison Experiment}
\label{sec:taishojikken}

As mentioned before, 
we use both the example rule 
and the semantic marker rule 
as judging rules. 
To check which rule is more effective, 
we made a comparison between 
the example method and 
the semantic marker method. 
The results are shown in Table \ref{tab:yourei_taishou}. 
The upper and lower rows of this table 
show the accuracy rates for training 
and test sentences, respectively. 
The precision of the method using examples 
was equivalent or superior 
to that of the method using semantic markers, 
as shown in Table \ref{tab:yourei_taishou}. 
This indicates that 
we can use examples as well as semantic markers. 
Since some codes in BGH are incorrect, 
we modified them. 
Since the precision using modified codes was higher than 
that using original codes, 
this indicates that 
the code modification is valid. 

\section{\normalsize \bf Summary}
\label{sec:owari}

In this paper, we presented a method of estimating 
referents of demonstrative pronouns, personal pronouns, 
and zero pronouns in Japanese sentences
using examples, surface expressions, topics and foci.
Unlike conventional works, 
which use semantic markers for semantic constraints, 
we use examples for semantic constraints 
and showed in our experiments that 
examples are as useful as semantic markers. 
We also proposed many new methods for estimating referents of pronouns. 
For example, we used the form ``X of Y'' for 
estimating referents of demonstrative adjectives. 
In addition to our new methods, 
we used many conventional methods. 
As a result, experiments using these methods 
obtained a precision rate of 87\% 
in estimating 
referents of demonstrative pronouns, personal pronouns, 
and zero pronouns 
for training sentences, 
and obtained a precision rate of 78\% 
for test sentences. 

{\footnotesize
\scriptsize

}

\end{document}